\documentclass[a4paper]{article}

\usepackage{INTERSPEECH2021}
\usepackage{CJKutf8}
\usepackage{color}

\title{BART based semantic correction for Mandarin automatic speech recognition system}
\name{Yun Zhao, Xuerui Yang, Jinchao Wang, Yongyu Gao, Chao Yan, Yuanfu Zhou}
\address{
  Cloudwalk Technology, Shanghai, China}
\email{\{zhaoyun, yangxuerui, wangjinchao, gaoyongyu, yanchao, zhouyuanfu\}@cloudwalk.cn}

\begin{document}
\begin{CJK}{UTF8}{gbsn}
\maketitle
\begin{abstract}
Although automatic speech recognition (ASR) systems achieved significantly improvements in recent years, spoken language recognition error occurs which can be easily spotted by human beings. Various language modeling techniques have been developed on post recognition tasks like semantic correction. In this paper, we propose a Transformer based semantic correction method with pretrained BART initialization, Experiments on 10000 hours Mandarin speech dataset show that character error rate (CER) can be effectively reduced by 21.7\% relatively compared to our baseline ASR system. Expert evaluation demonstrates that actual improvement of our model surpasses what CER indicates.  
\end{abstract}
	
\section{Introduction}

Achieving an acceptable word error rate (WER) is the key to success for any commercial ASR based products. In human computer interaction, ASR module lies in the front end and any transcription error can cascade in downstream tasks, such as intent recognition, named entity recognition and so on. Despite the fast development of ASR modeling technique and growing data resource, ASR systems still make mistakes which are obvious to humans, which degrades customer experience. Moreover, ASR systems are also prone to errors in presence of environment noise, microphone defects, speaker accent, speech disfluency etc. These persistent challenges motivated us to find an alternative approach to improve speech recognition accuracy. 

Language model is an essential component of ASR system. However, language knowledge is still not fully exploited for several reasons. In end-to-end ASR system, shadow fusion\cite{Zhao2019} is a widely used technique which incorporates external language models to improve recognition results. However, the left-to-right decoding procedure prohibits the language model from using right context and making corrections for long past histories. The correct path may be dropped during early steps. Hybrid systems typically incorporate a N-gram language model into a weighted finite state transducer (WFST) during the first pass decoding, and then use a deep neural network to search best path from word lattice generated from first step. Although deep neural models, such as RNN transformer, can utilize longer context and even bi-directional information, its searching space is limited to sequences constrained in word lattice. 

To overcome these problems, text-based correction using bidirectional language models is a strong candidate to further reducing CER. There are several similar text-based correction tasks, such as grammar correction \cite{Fu2018}, spelling correction \cite{Bassil2012} etc. However, they can’t be directly applied to ASR outputs. Grammar correction focuses on grammar integrity and is widely used in literal language. To contrast, semantic correction is mainly applied in spoken language. It corrects phrases which are correct in grammar but contradictory to common sense. Spelling correction models are typically trained to correct erroneous word which shares similar pronunciation or handwriting to correction target. However, the error distributions of their train datasets are significantly different from what ASR systems produce.

Recently, there has been a surge in developing text-based correction model for ASR systems \cite{Errattahi2018}. These researches can be generally classified into two categories: pipeline and sequence-to-sequence. In pipeline framework, error positions are firstly detected and multiple candidates with similar pronunciation are recalled at these positions. Then another module is used to rank these candidate sequences and outputs the best one. Text-based correction can also be solved in sequence-to-sequence fashion using encoder-decoder structure. This type of model takes raw ASR transcript in and outputs corrected transcripts directly. 

In this work, we proposed BART based semantic correction (SC) method which can effectively reduce CER of ASR systems. Our model is Transformer based. In training data preparation, we proposed a simple while efficient data expansion strategy to collect diverse samples with various ASR output errors. Meanwhile, pretrained language models which are trained on large corpus are introduced to initialize text correction model. In this way, universal language representation is transferred to semantic correction model. Lastly, separated input embedding layers are used in encoder and decoder side respectively to enhance representation for erroneous characters.

\section{Related Work}

There is a plethora of prior works on ASR systems output correction \cite{Zhang2019}\cite{Mani2020}\cite{Weng2020}. In this section, three lines of related work are briefly reviewed, pretrained language model, pipeline and end-to-end text-based correction method.

\subsection{Pretrained language model}
Unsupervised pretraining models on large corpus are capable of learning universal language representation. They have been shown to significantly improve most NLP downstream tasks by finetuning on relatively small task related dataset. It’s nature to incorporate external language information to a specific task. BERT\cite{Devlin2018} language model is based on Transformer encoder part. It is trained to recover original text from its copy with random masks and predict the order of two segments. In \cite{Liu2019}, a more robust pretraining strategy RoBERTa is proposed by removing next sentence prediction task in BERT and trained on a much larger corpus with longer iterations. BART language model shares the standard Transformer architecture. During pretraining, it aims to reconstruct original text from its noisy copy where insertion, deletion and substitution error may occur. BART has been shown to be particularly effective when finetuned for text generation tasks  \cite{Lewis2019}. The difference between BART pretraining objective and semantic correction task is minimal. Readers can refer \cite{Qiu2020} for a detailed review of more family of pretrained models.

\subsection{pipeline text-based correction method}

Generally, pipeline methods explicitly consist of three steps: error detection, candidates recall and reranking. In \cite{Bassil2012}, text errors were firstly detected by a N-gram model and a set of candidate words were suggested for substitution. In the end, a context-sensitive model is used to score all candidates and pick up the best one. In \cite{Zhang2020}, Bi-GRU network is used to assign error occurrence probability for each input token. Then vector representation of each token is interpolated with a special mask token representation according to those calculated probabilities. In the end, BERT model takes these noisy inputs and outputs corrected tokens.

\subsection{sequence-to-sequence text correction method}

In sequence-to-sequence text correction model, the semantics of input text with erroneous tokens is extracted and represented in a shared semantic vector space between encoder and decoder. Then decoder rephrases this semantic information using correct tokens. In \cite{Zhang2019}, Transformer-based text correction model is used to correct output of CTC-based Mandarin ASR systems. In \cite{Hrinchuk2019}, transfer learning technique is applied in training Transformer-based text correction models. However, university language knowledge is not fully transferred since Transformer model is initialized using only first 6 layers of pretrained BERT language model. To contrast, we used pretrained BART language model to initialize our text correction model, where all parameters of BART are transferred. 

\section{Our approach}

\begin{figure}[t]
  \centering
  \includegraphics[width=\linewidth]{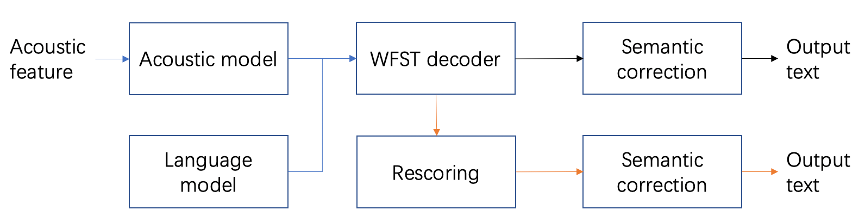}
  \caption{ASR system with semantic correction module.}
  \label{fig:fig1}
\end{figure}

Figure~\ref{fig:fig1} demonstrates the architecture of our proposed systems. Semantic correction can either be applied to 1st pass result or 2nd pass rescoring module to get further improvement, as shown in red path.  In order to sample diverse ASR errors, we investigated several effective techniques to extend the preliminary recognition results. Finally, the outputs generated by the decoder and reference texts are used to train a Transformer \cite{Weng2020} based semantic correction model.

\subsection{ASR baseline model}

Our baseline ASR system consists of three individual components: acoustic model, language model and a WFST-based decoder \cite{Mohri2002}. As to acoustic model, pyramidal Feedforward Sequential Memory Network (FSMN) \cite{Yang2018} is used. WFST searching graph is built from a 3-gram language model, lexicon, biphone convertor and HMM topology. RNN \cite{2010Recurrent}\cite{xu2018neural} and 4-gram language models are applied in rescoring word lattice generated during WFST decoding.

\subsection{Semantic correction model}

Standard 6-layer Transformer architecture is used in our semantic correction model. It consists of 6 encoder layers with input embedding layer, 6 decoder layers with output embedding layer, projection and Softmax layer. The modeling unit is in token level and word piece tokenizer \cite{Lewis2019} is used to split texts. Teacher forcing method is applied in training semantic correction model, where ASR hypothesis and reference texts are feed into input and output respectively. Beam search is adopted for decoding in deployment stage of semantic correction model. In this paper, parameter sharing strategy is experimented to find optimal representation for erroneous and correct tokens. We also investigated transfer learning efficiency by using RoBERTa and BART pretrained models. To make RoBERTa and BART model size comparable, hidden size in both configurations are set as 768. Other parameters are listed in ~\ref{tab:tab1}.

\begin{table}[t]
  \caption{Model parameters}
  \label{tab:tab1}
  \centering
  \begin{tabular}{p{0.16\columnwidth}p{0.11\columnwidth}p{0.14\columnwidth}p{0.09\columnwidth}p{0.09\columnwidth}p{0.09\columnwidth}}
    \toprule
    \textbf{Models} &\textbf{Encoder layers} &\textbf{Decoder layers} & \textbf{Hidden size} & \textbf{Heads} & \textbf{FF layer} \\
    \midrule
    Transformer & 6 & 6 &768 &12 & 3072  \\
    RoBERTa & 12 & 0 &768 &12 & 3072  \\
    BART & 6 & 6 &768 &12 & 3072  \\
    \bottomrule
  \end{tabular}
\end{table}

\subsection{evaluation}

\subsubsection{Character error rate}
The performance of Mandarin ASR system is usually measured by Character error rate (CER). In this paper, token is our counting unit instead of character. Hypothesis and reference text are firstly split into token sequences and then aligned by minimizing Levenshtein distance in between. Finally, CER is calculated as the sum of substitution, deletion and insertion errors divided by total tokens in reference sequence.

\subsubsection{Expert evaluation}
Transformer has an encoder-decoder architecture, which maps sequences in segment level semantic information, instead of individual tokens. To better evaluate the quality of our semantic correction model, complementary expert evaluation is necessary. First of all, part of errors can’t be corrected from text alone either due to lack of context or excessive errors in ASR outputs. Secondly, corrected sentence and reference sentence may share same semantic meaning but differs in expression. Thirdly, disfluent phrases in audios are transcribed inconsistently among different speech data resources. Correction on disfluency results in a higher CER value. Lastly, many foreign terms, especially named entities, have multiple equivalent Mandarin translations.

\section{Experiments}

\subsection{Data preparation}
Following experiments are conducted on a large Mandarin speech recognition task that consists of about 10,000 hours training data with about 8 million sentences. A test set contains about 5 hours data. In semantic correction model, the input and output take ASR hypothesis and its transcript respectively. Training and testing set are constructed by pairing of each hypothesis and its transcribed sentence. In our experiments, following techniques are adopted to create sample with diverse errors. 

\begin{enumerate}
\item Use a less robust acoustic model in decoding. In addition to our baseline model, we trained an another acoustic model using 10\% of full speech data with short epochs for SC training data generation.
\item Introduce error to acoustic features. We dithered MFCC features by multiplying a random coefficient ranging from 0.8 to 1.2. 
\item N-best expansion. During WFST decoding, we keep top 20 candidates and filter out sentences with CER at a proper threshold. This value is recommended to be less than 0.5 since original semantic meaning can hardly be recognized from ASR hypothesis with high error rate.
\end{enumerate}

\subsection{Baseline model training}

In our baseline ASR system, acoustic and language models are developed separately. Beam search method is applied in decoding graph. The acoustic model is trained on full speech dataset. 3gram, 4gram and RNN language models are built based on reference texts. All acoustic and language models are trained with Kaldi toolkits  \cite{Povey2011}.

\subsection{Embedding configuration}
Input and output texts share the same vocabulary in semantic correction task. However, erroneous tokens from input side convey more alternative meanings compared with their proper usage. Tokens in output side are expected to represent their canonical meanings. Therefore, it is beneficial to have separate representation at encoder embedding layer and decoder embedding layer. To test this hypothesis, we experiment with two different parameter sharing strategies between encoder and decoder embedding layer. Train dataset is expanded from full speech data and filtered by CER threshold at 0.3. Parameters are random initialized and updated by Adam optimizer during training. Beam search with beam width 3 is applied during inference. Hypothesis with highest score is selected as corrected text. 

In table~\ref{tab:tab2}, correction model with distinct input and output embedding reduces CER of our baseline output by 5.01\% relatively. To contrast, model with shared embedding deteriorates baseline recognition results. It demonstrates the necessity to use distinct representation in semantic correction tasks.

\begin{table}[t]
  \caption{ CER (\%) of ASR system with SC model under embedding configurations}
  \label{tab:tab2}
  \centering
  \begin{tabular}{llll}
    \toprule
    \textbf{3gram} &\textbf{+SC\_shared} &\textbf{+SC\_distinct}  \\
    \midrule
    7.78 & 7.91  & 7.39  \\
    \bottomrule
  \end{tabular}
\end{table}

\subsection{Transfer learning}
Next, we investigate transfer learning using different pretrained language models. Key parameters of RoBERTa and BART model are shown in Table 1. Both models were trained on 80GB Chinese corpus. Whole word masking technique was applied in RoBERTa pretraining \cite{Liu2019}\cite{Cui2019}. We followed \cite{Lewis2019} to train BART model. Initials weights of semantic correction model are transferred from pretrained language models. In BART based initialization, weights are directly applied to semantic correction model since they share the same encoder-decoder architecture. We followed \cite{Mani2020} to perform RoBERTa based initialization, where encoder and decoder part are both initialized with weights from first 6 layers of RoBERTa. Distinct representations for input and output are adopted in all following experiments. 

Compared with Random-initialized model, BART initialized model significantly pushes recognition quality forward, resulting in relative CER reduction by 21.7\% as shown in Table~\ref{tab:tab3}. Surprisingly, RoBERTa initialized model achieves marginally better results than Random initialized model in our test set. We suspect that universal language knowledge is not effectively transferred due to the gap between model architecture and training tasks. 

\begin{table}[t]
  \caption{CER (\%) of ASR system with SC model using different initialization}
  \label{tab:tab3}
  \centering
  \begin{tabular}{llll}
    \toprule
    \textbf{3gram} &\textbf{+SC\_RAND} &\textbf{+SC\_RoBERTa} &\textbf{+SC\_BART}  \\
    \midrule
    7.78 & 7.39  & 7.13 & 6.08 \\
    \bottomrule
  \end{tabular}
\end{table}

Our result also demonstrates that SC rivals with language model rescoring method in improving first pass decoding accuracy. In table~\ref{tab:tab4}, SC model achieves a much lower CER than 4gram rescoring. Although the CER of SC is slightly higher than RNN rescoring, their actually performance are close. Expect evaluation in later section shows that SC model is underestimated by CER metric. In addition, when SC is applied after LM rescoring module, performance can be further boosted. Table~\ref{tab:tab4} shows that CER is reduced by 21.1\% and 6.1\% relatively for 4gram and RNN respectively. 

\begin{table}[t]
  \caption{CER (\%)  of LM rescored output with SC model}
  \label{tab:tab4}
  \centering
  \begin{tabular}{lll}
    \toprule
    \textbf{rescore} &\textbf{baseline} & \textbf{+SC\_BART}  \\
    \midrule
    +4gram & 7.43  & 5.86  \\
    +RNN & 5.66 & 4.92 \\
    \bottomrule
  \end{tabular}
\end{table}

\subsection{Comparison with big language models}
In this experiment, we compare our BART based semantic correction model with big language models incorporated in ASR systems.  Big language models which are trained on a much larger corpus typically outperform small language models. However, their size is often limited for several practical reasons, such limited memory, decoding speed etc. Previous language models are trained on speech transcripts, which we refer to as small LMs. Here another set of big language models are built on an 8 times larger train set, which consists of abundant free available corpus in addition to speech transcripts.

When SC is applied on ASR system with small LM, its performance surpasses the same ASR system with large LM standalone. In Table~\ref{tab:tab5}, CER of former system is lower than the latter one by 5.4\% when using 3gram. Moreover, ASR system with large LMs can be further boosted by applying SC. CER of 3gram, 4gram rescoring and RNN rescoring are further reduced by 15.6\%, 14.4\% and 6.11\% respectively. It demonstrates that semantic correction model effectively learns universal language representation from BART pretrained model. A sample correction is shown in Table~\ref{tab:tab6}.
\begin{table*}[t]
	\caption{expert evaluation}
	\label{tab:tab7}
	\centering
	\begin{tabular}{p{0.4\columnwidth}p{0.2\columnwidth}p{1.35\columnwidth}}
		\toprule
		\textbf{Error type} &\textbf{Fraction} &\textbf{example} \\
		\midrule
		Insufficient context & 32.6$\%$ & 
			\begin{tabular}{p{1.35\columnwidth}}
		    ASR out    :看起来还行吧(\textcolor{red}{It} looks good) \\
		    Reference:\textcolor{red}{我}看起来还行吧 (\textcolor{red}{Do I} look good)\\
		    Correction: 看起来还行吧 (\textcolor{red}{It} looks good) \\ 
	        \end{tabular} \\
        \bottomrule
		Proper nouns error & 17.4$\%$ & 
			\begin{tabular}{p{1.35\columnwidth}}
				ASR out    :那当\textcolor{red}{克林}发现整件事都是个谎言的时候 (when \textcolor{red}{Colin} found out the whole thing was a lie)\\
				Reference:那当\textcolor{red}{柯林}发现整件事都是个谎言的时候 (when \textcolor{red}{Kolin} found out the whole thing was a lie)\\
				Correction: 那当\textcolor{red}{克林}发现整件事都是个谎言的时候 (when \textcolor{red}{Colin} found out the whole thing was a lie) \\
			\end{tabular} \\
       \bottomrule
			Equivalent expression & 23.3$\%$ & 
			\begin{tabular}{p{1.35\columnwidth}}
				ASR out    :几点\textcolor{red}{了} (\textcolor{red}{What} time is it)\\
				Reference:几点\textcolor{red}{了}  (\textcolor{red}{What} time is it)\\
				Correction: 几点\textcolor{red}{啦}  (\textcolor{red}{WHAT} time is it) \\
			\end{tabular} \\
       \bottomrule
			Corrector error & 26.7$\%$ & 
			\begin{tabular}{p{1.35\columnwidth}}
				ASR out    :\textcolor{red}{以襄陵}人注目的成就背后有太多的挫折（There are too many setbacks behind \textcolor{red}{the achievement of Xiangling people’s attention}）\\
				Reference:\textcolor{red}{一项令}人注目的成就背后有太多的挫折(There are too many setbacks behind \textcolor{red}{a remarkable achievement})\\
				Correction: \textcolor{red}{以相邻}人注目的成就背后有太多的挫折(There are too many setbacks behind \textcolor{red}{the achievement of neighbors’ attention})\\
			\end{tabular} \\
		
		\bottomrule
	\end{tabular}
\end{table*}

\begin{table*}[h]
	\caption{sample correction}
	\label{tab:tab6}
	\centering
	\begin{tabular}{p{0.92\columnwidth}p{0.92\columnwidth}}
		\toprule
		\textbf{ASR output} &\textbf{+ SC} \\
		\bottomrule
		\begin{tabular}{p{0.92\columnwidth}}
		    这种办法往往使\textcolor{red}{创立}越多的银行留成比例相对越小亏损越多银行留成比例相对越大\\ (This approach tends to make the proportion of banks with more \textcolor{red}{creation [Pinyin: chuangli]} smaller, and the proportion banks with more losses larger) \end{tabular}   & 
		\begin{tabular}{p{0.92\columnwidth}} 
		    这种办法往往使\textcolor{red}{创利}越多的银行留成比例相对越小亏损越多\textcolor{red}{的}银行留成比例相对越大\\ (This approach tends to make the proportion of banks with more \textcolor{red}{profits [Pinyin: chuangli]} smaller, and the proportion \textcolor{red}{of} banks with more losses larger)  \end{tabular}
	\end{tabular}
\end{table*}

\begin{table}[t]
  \caption{Performance of SC model on ASR system with large LM}
  \label{tab:tab5}
  \centering
  \begin{tabular}{p{0.2\columnwidth}p{0.2\columnwidth}p{0.2\columnwidth}p{0.2\columnwidth}}
    \toprule
    \textbf{LM type} & \textbf{Small LM + SC}  & \textbf{Large LM} & \textbf{Large LM + SC}  \\
    \midrule
    3gram & 6.08  & 6.43 & 5.42 \\
    +4gram & 5.86  & 6.20 & 5.31  \\
    +RNN &4.92 & 4.91 & 4.61 \\
    \bottomrule
  \end{tabular}
\end{table}

\subsection{Ablation studies}
We manually examined 300 samples where semantic corrector fails in correction. Most errors can be categorized into four types as shown in Table ~\ref{tab:tab7}. About one third of errors can’t be corrected from text feature alone. More context or acoustic information are needed. Proper nouns error constitutes 17\% of errors and most of them don’t affect semantic meaning or subsequent tasks such as name entity recognition. There are also 23\% of errors are misclassified by CER metric since semantic doesn’t change. Lastly, only 26.7\% of errors can be corrected from text alone while our model failed. Therefore, the overall correction quality is significantly higher than what CER metric indicate.

In addition, we experimented a data augmentation strategy using free corpus resources. Firstly, substitution distribution of each word is counted from real ASR output and reference pairs. Then augmented data are generated by randomly introducing errors to corpus based on this probability. This method can reduce CER by 8\% relatively when experiments are carried on 1/3 of speech data with 5 million augmented pairs. However, this improvement diminishes on full speech data.

\section{Conclusions}

In this work, we proposed an efficient transfer learning method in semantic correction model development for Mandarin ASR system. We demonstrated that it’s crucial to use distinct representation for input and output tokens. Experimental results show that BART initialized model significantly outperforms BERT initialized model. To further improve correction accuracy, we proposed a robust data augmentation strategy to adequately sample error distribution of ASR outputs. In addition, our proposed semantic correction has strong generalization ability. It systematically boosts the accuracy of ASR output regardless of upstream module. Lastly, our expert evaluation result shows that our model achieves more significant improvement than CER indicates. The persistent error motivates us to develop advanced modeling techniques utilizing additional features in future, such as acoustic feature, dialogue context, domain knowledge, etc.



\bibliography{mybib}
\bibliographystyle{IEEEtran}

\end{CJK}
\end{document}